# Bayesian Modeling Of An Human MMORPG Player

Gabriel Synnaeve[a,c] and Pierre Bessière[b,c]

[a]*Grenoble University*
[b]*CNRS*
[c]*E-Motion team INRIA Rhône-Alpes*
*655 avenue de l'Europe*
*38334 Montbonnot*

**Abstract.** This paper describes an application of Bayesian programming to the control of an autonomous avatar in a multiplayer role-playing game (the example is based on World of Warcraft). We model a particular task, which consists of choosing what to do and to select which target in a situation where allies and foes are present. We explain the model in Bayesian programming and show how we could learn the conditional probabilities from data gathered during human-played sessions.

**Keywords:** Bayesian programming, video games
**PACS:** 07.05.Mh

## INTRODUCTION

With more and more CPU cores and more and more immersive gameplays, AI is becoming a key feature of video games. Non-player characters (NPC) acting against the player as well as by his side are required to behave realistically and interestingly. We model a massively multiplayer online role-playing game (MMORPG) player through Bayesian programming (Lebeltel, 2004). We bet that this model can learn to play as a given player he could observe and that it will yield a realistic robot, customized to the instructing player.

A role playing game (RPG) consist in the incarnation by the human player of an avatar with a class (warrior, wizard, rogue, priest…) having different skills, spells, items, health points, stamina/energy/mana (magic energy) points. A MMORPG (e.g. World of Warcraft, AION or EVE Online) is a role-playing game in a persistent, multiplayer world. There are usually players-run factions fighting each other but we modeled a particular domain of multiplayer RPG called players versus environment (PVE). This is a cooperative task in which human players fight together against different NPC. And more specifically here, we modeled the "druid" class, which is complex because it can cast spells to deal damages or other negative effects as well as to heal allies or enhance their capacities ("buff" them).

This model deals only with a sub-task of a global AI for autonomous NPC. The problem that we try to solve with the presented model is: how do we choose which skill to use and on which target in a PVE battle? Possible targets are all our allies and foes. Possible skills are all that we know, we aim to get a distribution over target and skills and pick the most probable combination that is as yet possible to achieve (enough energy/mana, no cooldown). For that, we first choose what should be the target given all surrounding variables: is an ally near death that he should be healed, which foe should we focus our attacks on? Once we have the distribution over possible targets, we search the distribution on our skills, then we multiply by the one on targets. We put extra care in having the same input variables as a human player to keep consistent with our goal of modeling a human. However, some variables can be things that humans subconsciously interpolate from perceptions.

Approaches for handling NPC vary greatly and each has its advantages and drawbacks. Finite states machines (FSM), for instance used in Quake III and Warcraft III, lack a compact description, which causes analysis and control to be complex. Behavior trees (Isla, 2005), used in the Halo series and Left4Dead, address some of the analysis and design problems but are limited for collaborative behaviors without a "military" hierarchy. Killzone 2 (Champandard, 2009) ), in which robots ("bots") are a central element of the gameplay, uses hierarchical task networks (HTN) along with behavior trees. Behavior multi-queues (Cutumisu, 2009) resolve the problems of having collaborative, real-time and parallel behaviors. However, all of these approaches (often completed by specialized scripts) can attain human like behavior only with great efforts of the programmers to provide an exhaustive behaviors set. With the current and forthcoming games, it is a problem because the players (and the NPC) are able to do a lot of different actions. We try and invert this problem by having a model that includes machine learning techniques so that it can learn from experience and the programmers won't have to specify behaviors exhaustively. One of the smart applications of our model would be to leverage all the data that MMORPG servers collect from their players to automatically train NPC using Bayesian learning/techniques. As complexity of gaming worlds increases, we think that having a probabilistic approach increases robustness in the presence of incomplete information. Also, the "humanness" of a NPC is often considered as an important factor for the gameplay and the fun that the player will have, and perhaps even more in MMO in which a key point of the gameplay lies in interacting with humans (intelligent beings).

## BAYESIAN MODEL

### Bayesian Programming

Probability theory (Jaynes, 2003) is used as an alternative to classical logic to lead inference and learning as it is the only framework for handling inference in the presence of incompleteness and uncertainty. Bayesian programming (Diard, 2000 and Lebeltel, 2004) is a probabilistic framework encompassing the expressivity of Bayesian networks. The basic approach to work with Bayesian programming is first to translate incompleteness of the perceptible information into uncertainty that can then be handled by probability theory. The dependency between variables is specified as conditional probability, e.g. $A \Rightarrow B$ can be written as ***P(B=true | A=true) = 1*** . The structure of a Bayesian program is as follows:

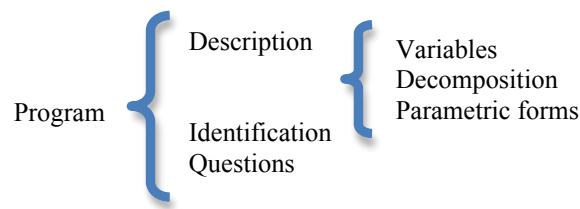

Our model will now be explained according to this structure. Once the joint decomposition, which consists in the *P(All Variables)*, is settled, one can ask questions to the Bayesian program that will be computed with the joint:

$$P(Searched \mid Known) = \frac{\sum_{Free} P(Searched \wedge Free \wedge Known)}{P(Known)}$$

which can be simplified by normalizing $P(Searched \mid Known) = \frac{1}{Z} \sum_{Free} JointDistribution$

## Description Of The Model

*Variables*

A very simple set of variables is as follows. We have the *n* characters as possible targets; each of them several bound variables. Health/Hit points (HP) are discretized in 10 levels, from the lower to the higher. Distance (D) is discretized in 4 zones around the robot character: contact where it can attack with a contact weapon and then 3 distances, the further one being "out of range even for the longest distance weapon/spell". Ally (A) is a Boolean variable mentioning if character *i* is allied with the robot character. Delta hit points (ΔHP) is a 3-valued interpolated value from the previous few seconds of fight that informs about the *i*th character getting wounded or healed (or nothing). Imminent death (ID) is also an interpolated value that encodes HP, delta HP and incoming attacks/attackers. This is a Boolean variable saying if the *i*th character if going to die anytime soon. This is an example of what we consider that an experienced human player will infer automatically from the screen and notifications. Class (C), simplified over 4 values, gives the type of the *i*th character: a Tank can take a lot of damages and taunt enemies, a Contact damager can deal huge amounts of damage with contact weapons (rogue, barbarian…), Ranged stands for the class that deals damages from far away (hunters, mages…) and Healers are classes that can heal (in considerable amounts). The Resist variable is the combination of binary variables of resistance to certain types of (magical) damages into one variable. With 3 possible resistances we get $2^3=8$ possible values. For instance "$R_i=FireNat$" means that the *i*th character resists fire and nature-based damages. Armor (physical damages) could have been included, and the variables could have been separated. The possible values of the skill variable are all the possible skills for the given character, and not only the available ones to cast at the moment to be able to have reusable probability tables (i.e. learnable tables).

Target: $T \in \{t_1 \ldots t_n\}$
Hit points: $HP_1 \ldots HP_n \mid HP_i \in [0 \ldots 9]$
Distance: $D_1 \ldots D_n \mid D_i \in \{Contact, Close, Far, VeryFar\}$
Ally: $A_1 \ldots A_n \mid A_i \in \{false, true\}$
Derivative hit points: $\Delta HP_1 \ldots \Delta HP_n \mid \Delta HP_i \in \{-, \circ, +\}$
Imminent death: $ID_1 \ldots ID_n \mid ID_i \in \{false, true\}$
Class: $C_1 \ldots C_n \mid C_i \in \{Tank, Contact, Ranged, Healer\}$
Resists: $R_1 \ldots R_n \mid R_i \in \{Nothing, Fire, Ice, Nature, FireIce, IceNat, FireNat, All\}$

Skill: $S \in \{Skill_1 \ldots Skill_m\}$

*Decomposition*

• <u>Target selection:</u> we want to compute the probability distribution on the variable Target (*T*), so we have to consider the joint distribution with all variables on which *T* is conditionally dependant : $T^{t-1}$ (the previous value of *T*), and all the variables on each character (except for Resists). The probability of a given target depends on the previous one (it encodes the previous decision and so all previous states). $HP_i$ depends on whether that the *i*th character is an ally, on his class, and if he's a target. Such a conditional probability table should be learned, but we can already foresee that a targeted ally with *C=tank* would have a high probability of having low HP because taking it for target means that we intend to heal him. $D_i$ is more probable to be far if $A_i$=*false* and *T=i* (our kind of druid attack with ranged spells). The probability of the *i*th character being an ally depends on if we target allies of foes more often. The probability that $\Delta HP_i$=*negative* is higher for $A_i$=*false* and $C_i$=*healer* and *T=i* and also for $A_i$=*true* and $C_i$=*tank*. As for $A_i$, the probability of $ID_i$ is driven by our soft evidence of targeting characters near death. The probability of $C_i$ is driven by the distribution of foes and allies population, tuned with a soft evidence of which classes our druid human player will target more frequently. Each and every time, if $T \neq i$, the probability of the left variable is given according to the uniform distribution. For the task of computing the distribution on Target, the joint distribution is simplified (by conditional independence of variables) as:

$$P(T \wedge T^{t-1} \wedge HP_{1:n} \wedge D_{1:n} \wedge A_{1:n} \wedge \Delta HP_{1:n} \wedge ID_{1:n} \wedge C_{1:n}) =$$
$$P(T^{t-1}).P(T|T^{t-1}).\prod_{i=1}^{n}\{P(HP_i|A_i \wedge C_i \wedge T).P(D_i|A_i \wedge T).P(A_i|T). \quad (1)$$
$$P(\Delta HP_i|A_i \wedge C_i \wedge T).P(ID_i|T).P(C_i|A_i \wedge T)\}$$

• <u>Skill selection:</u> As previously for targets, we are interested in the conditional probabilities of each character's state variables given other state variables and given *T* and *S*. If *T=i*, *S=big_heal*, $C_i$=*tank* and $A_i$=*true*, the probability that $HP_i$=0 or 1 (very low) is very high. Some skills have optimal ranges to be used at and so *P(Di)* will be affected. $A_i$=*true* will have a probability of *1.0* of *S=any_heal* as will $A_i$=*false* have a probability of *1.0* is *S=any_damage*. The probability of $\Delta HP_i$=*negative* will top when *S=heal* for an ally. That of $R_i$=*nature* for *S=nature_damage* will be very low. The probability of $ID_i$ will be high for *T=i* and *S=big_heal* or *S=big_damage* (depending on where *i* is an ally or not). For the task of computing the distribution on Skill we use:

$$P(S \wedge T \wedge HP_{1:n} \wedge D_{1:n} \wedge A_{1:n} \wedge \Delta HP_{1:n} \wedge ID_{1:n} \wedge C_{1:n} \wedge R_{1:n}) =$$
$$P(S).P(T).\prod_{i=1}^{n}\{P(HP_i|A_i \wedge C_i \wedge S \wedge T).P(D_i|A_i \wedge S \wedge T).P(A_i|S \wedge T). \quad (2)$$
$$P(\Delta HP_i|A_i \wedge S \wedge T).P(R_i|C_i \wedge S \wedge T).P(ID_i|S \wedge T).P(C_i|A_i \wedge S \wedge T)\}$$

*Parametric forms*

$P(T^{t-1})$ Unknown and so we chose it uniform.

$P(T|T^{t-1})$ Table, specified with a "prior" to prevent switching targets too often or simply learned. Uniform if there is no previous target.

$P(S)$ Unknown and so uniform, it could be a prior on *Skills* (not uniform).

$P(Left\_Value|Right\_Values)$ All others are learnable tables.

*Identification*

If there were only perceived variables, learning the right conditional probability tables would just be counting and averaging. However, some variables encode combinations of perceptions and passed states. We could learn such parameters through the EM algorithm but we propose something simpler for the moment as our "not directly observed variables" are not complex to compute, we compute them from perceptions as the same time as we learn. In the following Results part, we did not apply learning but instead manually specified the probability tables.

## Questions

In any case, we ask our model:

$P(S \wedge T | hp_{1:n} \wedge d_{1:n} \wedge a_{1:n} \wedge \Delta hp_{1:n} \wedge id_{1:n} \wedge c_{1:n} \wedge r_{1:n})$, which means that we want to know the distribution on *S* and *T* knowing all the state variables. We then choose to do the highest scoring combination of $S \wedge T$ that is available (skills may have cooldowns or cost more mana/energy that we have available).

Using (Bayes rule) *P(S,T)=P(S|T).P(T)*, to decompose this question, we can ask:

$P(T | hp_{1:n} \wedge d_{1:n} \wedge a_{1:n} \wedge \Delta hp_{1:n} \wedge id_{1:n} \wedge c_{1:n})$

which means that we want to know the distribution on *T* knowing all the relevant state variables, followed by (with the newly computed distribution on *T*)

$P(S | T \wedge hp_{1:n} \wedge d_{1:n} \wedge a_{1:n} \wedge \Delta hp_{1:n} \wedge id_{1:n} \wedge c_{1:n} \wedge r_{1:n})$

in which we use this distribution *T* to compute the distribution on *S* with:

$P(S = skill_1 | ...) = \left( \sum_T P(S = skill_1 | T \wedge ...).P(T) \right)$

We here choose to sum over all possible values of T. Note that we did not ask:

$P(S | T = most\_probable \wedge hp_{1:n} \wedge ...)$ but computed instead:

$\sum_T P(S | T \wedge hp_{1:n} \wedge d_{1:n} \wedge a_{1:n} \wedge \Delta hp_{1:n} \wedge id_{1:n} \wedge c_{1:n} \wedge r_{1:n})$

This computation has a high complexity (particularly when the sum has many terms, i.e. with a lot of targets), so we could choose not to do the sum and use and instantiate "most probable values", for instance of Target. Any such choice would lose information. There are possibly good combinations of *S* and *T* for a value of *T* that is not the most probable one. This downside may be so hard that we may want to reduce the complexity of computation by simplifying our model or its computation to be able to sum. We propose a solution in the discussion.

# Example

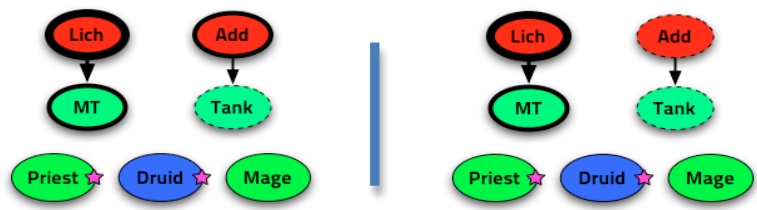

**FIGURE 1.** Example setup A (left) and B (right), 2 foes, 2 "tanks", players with stars can heal allies, players with dotted lines will soon die (*ID=true*).

This model has been applied to a simulated situation with 2 foes and 4 allies while our robot took the part of a "druid", a versatile class that can cast spells to do direct damages, damages over time, buff (enhancements), debuff, crowd-control, heal and heal over time. We display a schema of this situation in Figure 1. The arrows indicate foes attacks on allies. The larger the ring is, the more health points the characters have. MT stands for "main tank", Add for "additional foe". We worked with the skills corresponding to a Druid. *HOT* stands for heal over time, *DOT* for "damage over time", "abol" for abolition and "regen" for regeneration, a buff is an enhancement and a "dd" is a direct damage. "Root" is a spell which disables the target to move for a short period of time, useful to flee or to put some distance between the enemy and the druid to cast attack spells. "Small" spells are usually faster to cast than "big" spells.

*Skills* ∈ *{ small_heal, big_heal, HOT, poison_abol, malediction_abol, buff_armor, regen_mana, small_dd, big_dd, DOT, debuff_armor, root }*

We did not do the "Identification" part, which consists in learning the probability tables from observations. To keep things simple and because we wanted to explore the model, we worked with manually defined probability tables. So we introduced "soft evidences", parameters that will modify the conditional probability tables, which we will change to watch their effects. For instance the "soft evidence that a selected target is foe" and the "soft evidence that a selected target will soon die *(ID=true)*" that will consequently modify the probability tables of $P(A_i)$ and $P(ID_i)$ respectively. We set the probability to target the same target as before to 0.4 and the previous target to *Lich* so that the prior probability for all other 6 targets is 0.1 (4 times more chances to target the *Lich* than any other character). We set the soft evidence $P(A_i=false|T=i)$ to 0.6. This means that our robotic *Druid* is mainly a damage dealer and not a healer. For the "target selection" model, we can see on Figure 2.A that the evolution from selecting the main foe *Lich* to selecting the ally *Tank* is driven by the increase of "soft evidence that a selected target will soon die" and our robot eventually moves on targeting his *Tank* ally (to heal him). We can see on Figure 2.B that, at some point, the robotic *Druid* prefers to kill the dying *Add* to save his ally *Tank* instead of healing him. Note that there is no variable showing the relation between *Add* and *Tank* (the first is attacking the second, who is taking damages from the first), but this is under consideration for a future, more complete, model.

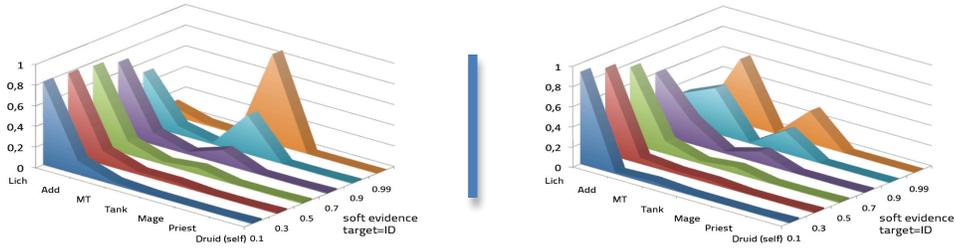

**FIGURE 2.A.** (left) Probabilities of targets depending on the soft evidence that a target is dying with setup A | **FIGURE 2.B.** (right) Same, with setup B

For the "skill selection" model, we can see on Figure 3 the influence of $ID_i$ on Skill which is coherent with the Target distribution: either, in setup A, we evolve with the increase of $P(ID_i=true|Target=i)$ to choose to heal our ally or, in setup B, to deal direct damage (and hopefully, kill) the foe attacking him. As you can see here, when we have the highest probability to attack the main enemy ("Lich", when $P(ID_i=true|Target=i)$ is low), who is a $C=tank$, we get a high probability for the Skill *debuff_armor*. We only cast this skill if the debuff is not already present, so perhaps that we will cast *small_dd* instead. To conclude this example, Figure 4 shows the distribution on $P(T,S|all\_status\_variables)$ with setup A and a the probability to target the previous target (set to *Lich* here) only ~2 times greater than any other character (so that we focus less on the same character), soft evidences $P(ID_i=true|Target=i)=0.9$ and $P(A_i=false|Target=i)=0.6$. In a greedy way, if the first couple (*T,S*) is already done or not available, we take the second.

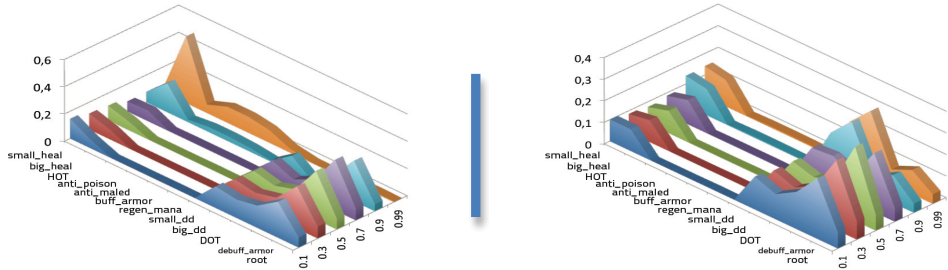

**FIGURE 3.A.** (left) Probabilities of skills depending on the soft evidence that a target is dying with setup A | **FIGURE 3.B** (right) Same, with setup B

# DISCUSSION

## Perspectives

These results are encouraging, however the most challenging aspect will be to complete the model for different kind of tasks and for the robot to be able to learn sequences of actions, by developing forecasting strategy. This model has to be applied in a real MMORPG, out of its simulation, to reveal all its shortcomings and be improved. We foresee some future difficulties, for instance there is a possibility for many games that the Skill variable will be very big and that it will imply a too high computational cost. For that concern, we propose to cluster the skills in global skills (*GS*). This approach to break down the complexity of computation is *general* and can be used with other variables.

# Conclusion

We believes that modeling the human behavior has two big advantages of being able to learn from human-played examples so that it can generate a behavior (control the character) as well as predict what will the other players do. Learning from game sequences with a human model allows for fast development of realistic AI in video games. It also decreases the complexity from having many possible actions and possible behaviors thanks to the behavior learning part. Applying the model to other players for prediction enables the robot to have these predictions as input variables, and act accordingly ("I think that you think"...). Human players do long term planning, consciously or not. However, modeling high-level cognition is very hard. We assume that by modeling a reactive robot with regard to observable variables and predicted variables, we can emulate human planning. In particular, some variables are future extrapolated values, some others encode past states and, in the future, we plan to have variables corresponding to more elaborated predicted states through other Bayesian programs.

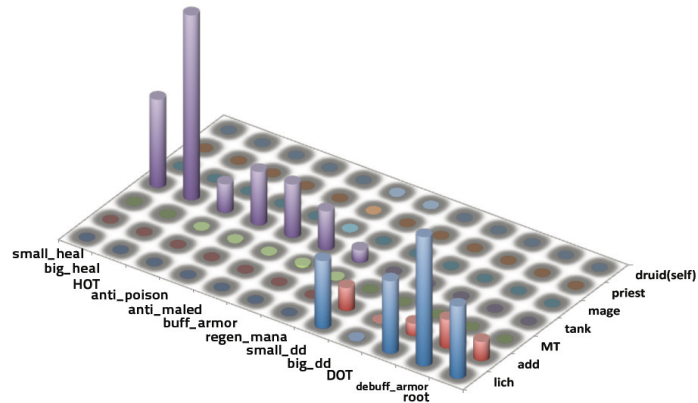

**FIGURE 4.** Log-probabilities of Target and Skill with setup A, *P(ID|Target)=0.9, P(A|Target)=0.6*